\documentclass{article}

\usepackage{arxiv}

\usepackage[utf8]{inputenc} 
\usepackage[T1]{fontenc}    
\usepackage{hyperref}       
\usepackage{url}            
\usepackage{booktabs}       
\usepackage{amsfonts}       
\usepackage{nicefrac}       
\usepackage{microtype}      
\usepackage{lipsum}		
\usepackage{graphicx}
\usepackage{doi}
\usepackage{amsmath,amssymb}
\usepackage{comment}
\usepackage{caption}
\usepackage{cite}
\usepackage{subcaption}
\usepackage{authblk}
\usepackage[english]{babel}
\usepackage{svg}
\usepackage{csquotes}
\usepackage{pgfplots}
\pgfplotsset{compat=1.18}
\usetikzlibrary{fit}
\usepackage{float}
\usepackage{xcolor}
\definecolor{darkgreen}{rgb}{0.0, 0.5, 0.0} 
\usepackage[authoryear]{natbib} 
\usepackage{xcolor}
\definecolor{teal_blue}{HTML}{003B46} 
\definecolor{ash_gray}{HTML}{BDC3C7} 
\definecolor{burgundy}{HTML}{6A1B4D}
\definecolor{mustard}{HTML}{E1AD01}

\usepgfplotslibrary{colorbrewer}
\usetikzlibrary{pgfplots.statistics, pgfplots.colorbrewer} 
\usepackage{pgfplotstable}
\usepackage{filecontents}

\usepackage{subcaption}
\usepackage{caption}
\usepackage{xcolor}

\title{Variational Rank Reduction Autoencoders for Generative Thermal Design }

\renewcommand{\headeright}{Technical Report}
\renewcommand{\shorttitle}{Rank Reduction Autoencoders for Generative}

\hypersetup{
pdftitle={Variational Rank Reduction Autoencoders for Generative},
pdfsubject={q-bio.NC, q-bio.QM},
pdfauthor={A. Tierz},
pdfkeywords={First keyword, Second keyword, More},
}
\author[1]{Alicia Tierz}
\author[2]{Jad Mounayer}
\author[2]{Beatriz Moya }
\author[2,3]{Francisco Chinesta}

\affil[1]{{\small ESI Group-UZ Chair of the National Strategy on Artificial Intelligence. \protect\\ Aragon Institute of Engineering Research (I3A). Universidad de Zaragoza. Zaragoza, Spain.}}
\affil[2]{{\small PIMM Lab. ENSAM Institute of Technology. Paris, France. }}
\affil[3]{{\small CNRS@CREATE LTD. Singapore.}}

\begin{document}
\maketitle

\begin{abstract}
Generative thermal design for complex geometries is fundamental in many areas of engineering, yet it faces two main challenges: the high computational cost of high-fidelity simulations and the limitations of conventional generative models. Approaches such as autoencoders (AEs) and variational autoencoders (VAEs) often produce unstructured latent spaces with discontinuities, which restricts their capacity to explore designs and generate physically consistent solutions.  

To address these limitations, we propose a hybrid framework that combines Variational Rank-Reduction Autoencoders (VRRAEs) with Deep Operator Networks (DeepONets). The VRRAE introduces a truncated SVD within the latent space, leading to continuous, interpretable, and well-structured representations that mitigate posterior collapse and improve geometric reconstruction. The DeepONet then exploits this compact latent encoding in its branch network, together with spatial coordinates in the trunk network, to predict temperature gradients efficiently and accurately.  

This hybrid approach not only enhances the quality of generated geometries and the accuracy of gradient prediction, but also provides a substantial advantage in inference efficiency compared to traditional numerical solvers. Overall, the study underscores the importance of structured latent representations for operator learning and highlights the potential of combining generative models and operator networks in thermal design and broader engineering applications.

\end{abstract}

\section{Introduction}

Fast solvers and surrogates are essential for resolving temperature fields over complex geometries in different fields of engineering and scales. These applications range from mechanical elements and their manufacturing process \cite{chen2022data,sajadi2024real,pan2024temperature} to urban analysis \cite{huo2022study,kim2022estimating,tanoori2024machine} to ensure the performance, durability, and safety of the system.

However, high-fidelity thermal simulations in practical applications, such as telecommunications and manufacturing, are often computationally expensive. Traditional solvers predominantly employ finite-element or finite-volume methods to predict thermal behaviour \cite{song20003,CHAQUET2025104385}. These methods are computationally intensive and time-consuming, especially when implemented in large-scale systems, real-time applications, or iterative design processes. 

The use of deep learning solvers results also inefficient since the design space often comprises thousands or more parameters \cite{zhou2024evolutionary,Shan2010}. Consequently, reducing the dimensionality of the design space is essential for both simulation and generative optimisation. In this case, a major challenge in using deep learning solvers with reduction techniques is to effectively parameterize the design domain in the new manifold \cite{kashefi2022physics,HE2024117130}. 

The development of surrogate models grounded in reduced-order models and machine learning, particularly deep learning, has been extensively investigated in recent years \cite{su2021super,zhang2025study}. Having surmounted the initial exploratory stages in black-box models, numerous alternatives in physics-informed machine learning are now available \cite{chen2025physics,toscano2024inferring}. In this context, deep operator networks (DeepONets \cite{Lu_2021}) have emerged as an innovative and promising architecture for learning nonlinear operators directly from data. Their theoretical foundation is based on the Universal Approximation Theorem for Operators, which states that a neural network, even with a single hidden layer, can accurately approximate any continuous nonlinear operator. However, these solvers have not yet been comprehensively integrated with generative tools and spaces to thoroughly investigate solutions in optimisation and inverse problems. 

Generative artificial intelligence methodologies, particularly autoencoders, variational autoencoders, and generative adversarial neural networks, are employed to learn spaces, or manifolds. These methodologies facilitate the embedding of high-dimensional design spaces derived from high-dimensional parameter spaces, or even from images or NURBS, with the objective of determining a reduced-order parameterization \cite{kang2024physically,wang2025generative}. This reduced parameterization serves as a foundational tool for the exploration and analysis of the design space, including inverse analysis \cite{brzin2024using,kang2024physically}. Nevertheless, these methodologies, rooted in deep neural networks, notwithstanding their preliminary reconstruction metrics, yield discontinuous solution spaces. This phenomenon, often referred to as ``holes'', pertains to regions within the manifold that, due to various factors, primarily data scarcity, fail to be reconstructed with precision and coherence. Consequently, the outputs within these regions are devoid of significance and fail to accurately represent the underlying design space \cite{bengio2013representation}. Furthermore, traditional Variational Autoencoders (VAEs), although they address continuity through probabilistic modelling, can suffer from posterior collapse and tend to generate blurred images because KL divergence is their only form of regularisation.

This study presents a thermal solver for shapes inspired by manufacturing and electronic systems. The solution space is developed upon a linearly interpolable manifold, which serves as an efficient generative tool seamlessly integrated with the simulation process. Analyzing these temperature distributions enables the optimization of device design to mitigate excessive heat, which can otherwise lead to performance degradation, or even mechanical failure. The foundation of this approach lies in the use of rank reduction autoencoders (RRAEs) \cite{mounayer2025rankreductionautoencoders}, with a particular emphasis on their variational approximation (VRRAEs)\cite{mounayer2025variational}. This methodology incorporates singular value decomposition (SVD) in conjunction with traditional autoencoders to enhance the quality of the reduced manifolds. As a result, the new reduced parametrization can be coupled with solvers, in this case DeepONets, to perform physics prediction of the designs from the reduced exploration space.

Section 2 expands the state-of-art and challenges of generative AI  and surrogate modeling. Section 3 states the methodology with a detailed description of the rank reduction autoencoder used and its combination with surrogates to predict the temperature fields of the geometries. Results are presented in section 4, demostrating the outperformance of the proposed method with respect to other state-of-art strategies. To finalize, we present the conclusions and future perspectives of this work for the further development and use of the presented approach.

\section{Related Work}

\subsection*{Generative Artificial Intelligence}

Generative Artificial Intelligence (Gen-AI) constitutes a significant challenge within the current landscape of artificial intelligence research, and particularly in engineering applications \cite{regenwetter2022deep}. It can be argued that the foundation of these methodologies is rooted in the techniques associated with reduced order modeling, from POD and non-linear manifold learning to autoencoders (AE) \cite{loaiza2024deep}. In essence, these techniques learn from data the spaces where the problem object of study is embedded, thereby facilitating the exploration of the manifold to identify further potential solutions regarding a new objective. Autoencoders are among the most utilized tools due to their superior performance in the presence of nonlinear correlations within data. However, vanilla autoencoders struggle in learning effective representations in a complex setting, which can often result in a latent space suffering from ``holes'', i.e. areas where the corresponding manifold is undefined, and its prediction does not produce physically meaningful predictions \cite{bengio2013representation}.

There exist two main methodologies that attempt to address this challenge. Variational Autoencoders (VAEs) \cite{kingma2013auto} extend the concept of traditional autoencoders by applying Bayes’ theory. With this constraint, the AE does not learn a deterministic space, but a distribution that provides continuity and smoothness to the manifold of solutions. 
This technique has been applied in a variety of fields.
The authors in \citep{wang2025generative} exploit the capabilities of VAEs for aerofoil design from NURB entries. Similarly, Guo et al. \citep{guo2018indirect} apply this technique to topology optimisation for heat conduction design from images. Yang et al. \citep{yang2024research} use $\beta$-VAEs to enhance the interpretability of the manifold in multi heat source designs of electronic devices, applying a parametrization of the domain to perform forward analysis. Generally, the efficacy of VAEs is heavily reliant on the imposition of suitable priors, and it exhibits limited variability in the generated samples if the distribution is not effectively learnt. Authors in \citep{casale2018gaussian}, for instance, propose an approach to learn an extended approximation of distributions by employing Gaussian processes (GP) in the training process.

To expand the generative capabilities of autoencoders, Generative Adversarial Neural Networks (GANs) provide a new paradigm of learning that trains with samples beyond the provided dataset \cite{goodfellow2014generative}. GANs learn through a training game where two networks are challenged; a generator learns to produce good samples that trick a discriminator, while the discriminator progresses to be able to distinguish fake from real samples more efficiently. As a result, the network learns an enriched latent manifold. The work of Qian et al. \citep{qian2022adaptive} uses this technique to learn different component layouts and performs optimisation of the designs through the use of genetic algorithms \citep{chelouah2000continuous} to explore the solution manifold. In this case, a convolutional neural network is used as a surrogate to predict the heat field of the designs to accelerate the optimization process. 

Often, GANs and VAEs tend to be combined to merge the capacities of both tools, especially in the need to obtain a forward connection with the GAN generative manifold \cite{bermejo2024thermodynamics}. It is also worth noting that, even though GANs show a great performance in design sampling with highly realistic results, manifolds of solutions of physical information such as dynamic data were better learnt with VAEs than GANs, thus demonstrating better performance for data assimilation and prediction \cite{bao2022variational}. 

In spite of partially solving the issues presented in the generative capabilities of the AE, VAEs and GANs still present problems at the time of having consistent results in the interpolation scheme. In other words: these techniques lack compatibility and interpretability constraints to learn regularized and interpolable spaces which can represent the real variability of a problem.

Learning interpretable manifolds leads to better performant spaces. That is the case of reduced spaces conditioned by physics biases \cite{takeishi2021physics,bacsa2023symplectic,zhong2023pi,yonekura2025generating}, or those where additional restrictions are imposed \cite{solera2024beta}, such as topology \cite{moor2020topological} or orthogonality \cite{eivazi2022towards}, inspired by the properties of POD in the interpretation of manifolds.

This study proposes a methodology based on the use of Rank Reduction-type autoencoders \cite{mounayer2025variational}, which present an architecture that couples autoencoders (AEs) with singular value decomposition (SVD). This integration leverages the compression and pattern extraction capabilities inherent in deep learning. Then, it employs SVD to ensure that the resultant reduced space exhibits interpretability, continuity, and linearity.

\subsection*{Physics solvers applied to reduced manifolds}

Several works in the literature tackle the resolution of PDEs \cite{fresca2022pod,fresca2021comprehensive,pichi2024graph,wang2024latent,mavi2023unsupervised} and complex, usually non-linear, dynamical systems \cite{suman2022investigation,maulik2021reduced,bakarji2023discovering} through the use of model order reduction, also including some type of bias to exploit the benefits of physics-informed machine learning \cite{bermejo2024thermodynamics,he2025thermodynamically}. In the aforementioned works, deep learning techniques and several leading autoencoder architectures are employed to learn a significant, often compressed, representation of the physical patterns to be predicted. 

Zong et al., for instance, extract geometrically representative descriptors in a new manifold to apply Neural Operators \cite{zhong2025physics}. Also on the use of neural operators, Latent Neural Operators use the so-called Physics-Cross-Attension (PhCA) block acting as the encoder and decoder to obtain an interpretable approximation of neural operators in the reduced space \cite{wang2024latent}. It is also worth noting the works on the use of the GENERIC formalism as an approximation of the dynamics based on thermodynamic principles, that can be easily transferred to reduced manifolds since evolution of energy and entropy can be also expressed with the new coordinates \cite{bermejo2024thermodynamics, hernandez2021deep, moya2022physics, he2025thermodynamically}. Regarding PINNs and DeepONets, authors in \cite{sholokhov2023physics} employ numerical analysis collocation methods to integrate information derived from a known equation into the latent-space dynamics of a reduced-order model (ROM). Inspired by Gen-AI, Taufik and Alkhalifah employ latent diffusion models as an alternative to VAEs or GANs due to their high performance on information reconstruction to learn compressed latent representations of the distribution of PDEs \cite{taufik2025latentpinns}. In the case of \cite{oommen2022learning}, the authors explore in this work the use of DeepONets in the new manifold reached with autoencoders.

Nevertheless, as articulated in the preceding section, the absence of imposed biases during the learning process may compromise the quality and continuity of the manifold, as well as the expressiveness of the latent variables in terms of parameterisation capabilities, thereby adversely impacting the performance of the physics surrogate. The pursuit of interpretability, defined as the capacity to comprehend the internal mechanics of the model  within manifolds \cite{cicirello2024physics} has led to a series of studies that embed knowledge into the new spaces \cite{liu2022physics,takeishi2021physics,lu2020extracting,pakravan2021solving,taufik2025latentpinns,lu2022discovering}. 

Proper Orthogonal Decomposition (POD) is typically regarded as more interpretable than autoencoders. In POD, each mode possesses a distinct physical significance, frequently corresponding to dominant patterns or structures within the system, such as flow vortices or deformation modes, whereas autoencoders embody a more abstract representation of the space. Furthermore, POD is derived from the process of eigenvalue decomposition or singular value decomposition (SVD) applied to the covariance matrix. This allows for the ranking of modes according to their energy content, thereby offering a clear depiction of the degree to which each mode embodies the system's behavior. In contrast, autoencoders necessitate training via backpropagation, and the resulting weights are not readily interpretable. Without explicit enforcement, such as through regularization or constraints, there is no inherent conception of captured energy or variance.

POD has been deeply used in the study of complex systems. Nonetheless, it also has impacted the deep learning perspective. Authors in \cite{baker2023learning} demonstrate that neural ODEs exhibit enhanced performance when trained using the Proper Orthogonal Decomposition (POD) modes rather than the original spatial coordinates of the physical system, a methodology similarly suggested in \cite{sentz2025reduced}. As stated previously, the reason lies in the emergence of patterns and yet physical significance compared with other model order reduction techniques. Following this idea, Eivazi et al. learn DeepONets from a representation learnt with kernel Principal Component Analysis (kPCA), where the manifold also benefits from the benefits of the SVD, claiming outperformance of the method compared to ordinary DeepONets \cite{eivazi2024nonlinear}.

In this work, we propose the use of a DeepOnet to learn the operators that lead to the prediction of the temperature field of each design from the parametrization learnt with Variational Rank Reduction-type autoencoders\cite{mounayer2025variational}. Our justification for employing the DeepOnet is based on our awareness on the presence of correlations that can be discerned through operator learning. This forms the basis for developing an interpretable surrogate model, drawing inspiration from physics-informed deep learning methodologies.

\section{Methodology}
We propose a hybrid learning framework that combines a Variational Rank Reduction Autoencoder (VRRAE) \cite{mounayer2025variational} and a Deep Operator Network (DeepONet)\cite{Lu_2021} for fast and accurate prediction of temperature gradients in plates with varying internal geometries. This approach aims to replace traditional numerical solvers for the heat equation with a data-driven surrogate that generalises across different domain configurations. The complete architecture is represented in figure \ref{fig:net_structure}.

\subsection{Problem fomulation}

We address the problem of learning the mapping from a 2D geometry to its corresponding temperature gradient field. Geometries are represented as images that encode internal structures (e.g., holes or inserts) embedded within a rectangular plate. These geometries vary across samples, adding complexity to the solution space.

In this description, $u \in \mathcal{U}$ stands for a geometry defined in a continuous domain, 
where $\mathcal{U}$ represents the space of geometries. Each geometry is represented by a two-dimensional image given by 
\[
\mathbf{X} = \mathcal{I}(u), \quad \mathbf{X} \in \mathbb{R}^{m \times n},
\]
where $\mathcal{I} : \mathcal{U} \rightarrow \mathbb{R}^{m \times n}$ 
is the operator that maps a geometry to its pixelized form on a grid of size $m \times n$.

The goal is to predict the temperature gradient $\nabla T(\mathbf{x})$ 
at any spatial location $\mathbf{x}(x,y) \in \Omega(u) \subset \mathbb{R}^2$, 
where $\Omega(u)$ is the spatial domain associated with a geometry $u$. 
Ground-truth solutions are obtained by numerically solving the heat equation 
using a finite element solver. 
Each data sample is represented as a triplet $(u, \mathbf{x}, \nabla T(u,\mathbf{x}))$, 
which enables the use of operator learning frameworks.

This work proposes an encoding of $X$ into a generative space to find a mapping from the pixelized representation of the original design $u$ to the gradient $\nabla T(u,\mathbf{x})$.

\subsection{Latent Geometry Encoding via VRRAE}

To encode the geometric variability of the input domain in a compact and meaningful form, we employ a Variational Rank Reduction Autoencoder (VRRAE). This architecture combines the probabilistic modeling of Variational Autoencoders (VAEs) with the structured bottleneck of Rank Reduction Autoencoders (RRAEs), making it particularly suited for problems where the input function (here, the plate geometry) exhibits local or low-rank structure.

The VRRAE comprises two main components:
\begin{enumerate}
    \item An encoder network that maps the input geometry (a 2D image) to a latent representation.
    \item A decoder network that reconstructs the original image from a sampled version of this latent representation.
\end{enumerate}

\subsubsection*{Latent structure via truncated SVD}
Unlike conventional VAEs, the VRRAE introduces a truncated Singular Value Decomposition (SVD) within the latent space. The encoder maps each input image  to a latent vector. For a batch of $N$ input samples, the encoder produces a laten matriz $\mathbf{Y} \in \mathbb{R}^{L\times N}$, where $L$ is the dimension of the latent space before truncation. This matrix $\mathbf{Y} $  is then factorized using SVD as $\mathbf{Y}  = \mathbf{U}\mathbf{S}\mathbf{V}^T$. The representation is then truncated to a fixed rank $k^*$, yielding a compressed latent code $\mathbf{Y}  = \bar{\mathbf{U}}\bar{\mathbf{S}}\bar{\mathbf{V}}^T$, where only the top $k^*$ singular components are retained. In our application, this bottleneck rank is set to $k^*=8$, effectively reducing the high-dimensional geometry to an 8-dimensional vector.

This structure not only enforces a natural ordering of importance in the latent dimensions (via the singular values), but also introduces a strong form of regularization, as it restricts the model to reconstruct images only from their most dominant modes. Another advantage is that it is imposed without adding additional terms to the loss function during optimisation.

\subsubsection*{Variational modeling and mean function choice}

To preserve the benefits of the variational framework, the VRRAE introduces stochasticity by sampling the singular values (or their coefficients $\tilde{\alpha}_{i,j}$) from a conditional distribution. Specifically, instead of directly using the deterministic coefficients of the truncated SVD ($\bar{\alpha}_{i,j}$),  the model samples coefficients $\tilde{\alpha}_{i,j}$ from a distribution $q(\tilde{\alpha}_{i,j}|\mathbf{X})$, typically modeled as Gaussian. 


In the VRRAE formulation, the mean of this distribution is fixed to the deterministic SVD coefficients. Equivalently, the expected values $\mathbb{E}(\tilde{\alpha}_{i,j}) = \bar{\alpha}_{i,j}$, $\forall i,j$, coincide with the coefficients obtained from the SVD of $Y$, while a neural network predicts the corresponding standard deviations.

This choice is crucial for preserving the structured regularization properties of RRAEs within the variational framework. By aligning the mean of the latent distribution with the truncated SVD, the model retains the interpretability and ordering of the latent dimensions, while still benefiting from the flexibility and robustness provided by variational inference.

Moreover, this construction ensures that the orthonormality of the basis vectors 
$\mathbf{U}_k$ is preserved during training, and it limits the risk of posterior collapse, a common issue in VAEs where the latent variables lose their dependence on the input. In the VRRAE, collapse can only occur toward structured, bounded values, which maintains informative representations even under regularization pressure.

\begin{figure}
  \centering
  \includegraphics[width=\textwidth]{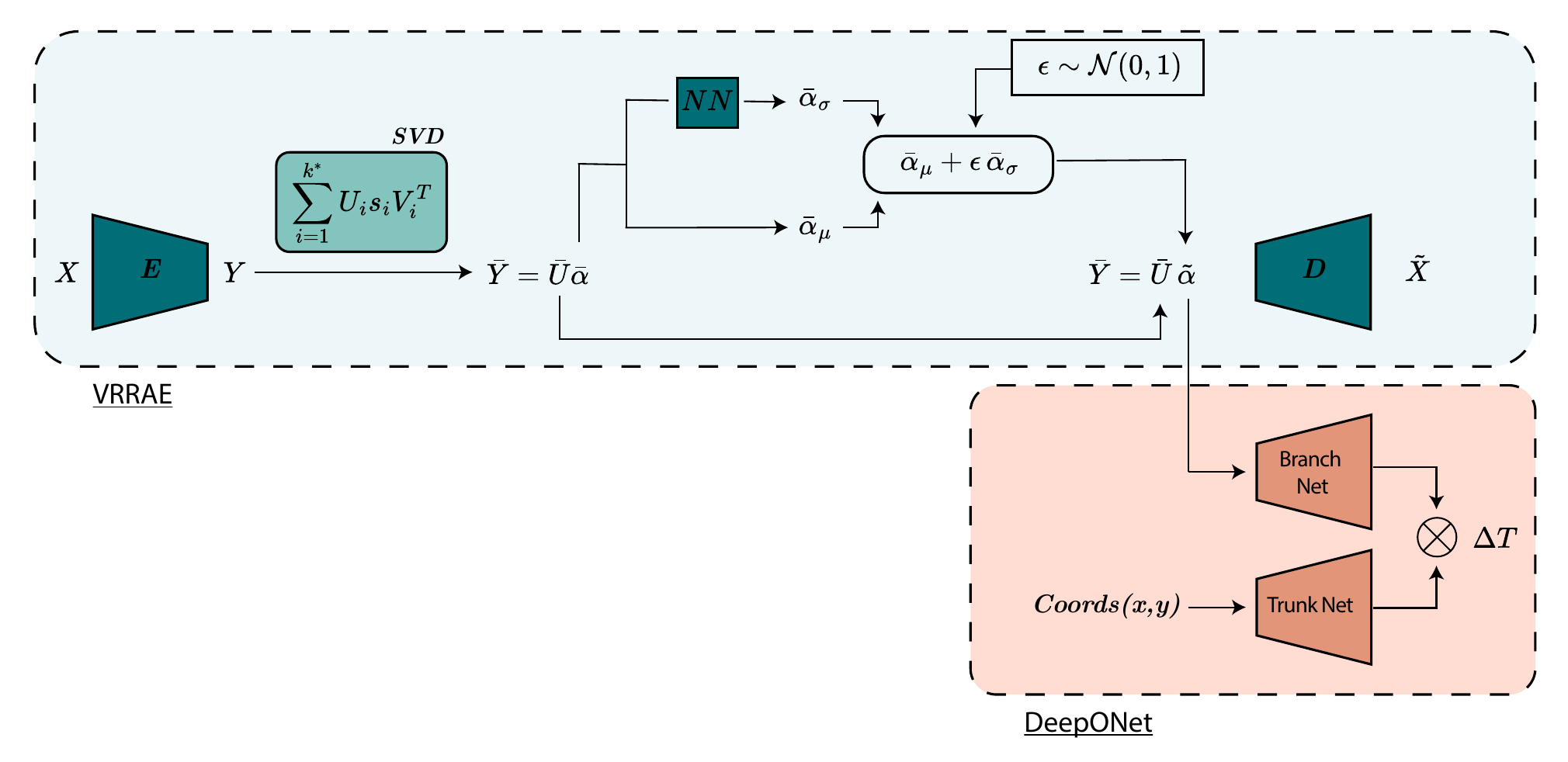}
  \caption{Schematic illustrating the architecture combining a Variational Rank Reduction Autoencoder and a DeepONet.}
  \label{fig:net_structure}
\end{figure}

\subsubsection*{Training objective} 
The VRRAE is trained by minimizing a composite loss:
\begin{equation}\label{eqn:loss_vrrae}
    \mathcal{L}_{VRRAE}=\underbrace{\|\mathbf{X}-\tilde{\mathbf{X}}\|_2}_{\mathcal{L}_{rec}} + \beta\underbrace{\frac{0.5}{N}\,\,\text{sum}(\mathbf{1}_{k^*\times N}+\log(\bar{\mathbf{\alpha}}_\sigma\odot\bar{\mathbf{\alpha}}_\sigma)-\bar{\mathbf{\alpha}}_\mu\odot \bar{\mathbf{\alpha}}_\mu-\bar{\mathbf{\alpha}}_\sigma\odot\bar{\mathbf{\alpha}}_\sigma)}_{\mathcal{L}_{KL}},
\end{equation} 

where $\mathcal{L}_{rec}$ is the reconstruction loss, measured as the Mean Squared Error (MSE) between the original and reconstructed geometry images and $\mathcal{L}_{KL}$ is the Kullback-Leibler divergence between the posterior distribution $q(\tilde{\mathbf{\alpha}}|\mathbf{\mathbf{X}})$ and a standard Gaussian prior. This term encourages the sampled coefficients to remain within a structured and bounded region of the latent space. The hyperparameter $\beta$ controls the strength of the KL regularization and is selected through empirical tuning.

\subsection{Physics prediction based on DeepONets}

Following the latent geometry encoding provided by the VRRAE, the core predictive model is a Deep Operator Network (DeepONet), a neural architecture designed to approximate nonlinear operators, i.e., mappings between function spaces. DeepONets are theoretically grounded in the Universal Approximation Theorem for Operators, which guarantees their capacity to learn continuous operators under mild assumptions.

A DeepONet consists of two sub-networks:  A branch net, which processes the input function, and a trunk net, which processes the coordinates where the output is evaluated.
The final output is computed as the inner product between the outputs of these two sub-networks:
\begin{equation}\label{eqn:innerproduct}
G(\tilde{\alpha} )(\mathbf{x}) \simeq  \sum_{k=1}^{p} b_k (\tilde{\alpha} )t_k(\mathbf{x})
\end{equation}

where $G$ is the operator approximating the temperature fields, $b_k$ are the outputs of the branch net (associated with the input $\tilde{\alpha}$ ) and $t_k$ those of the trunk net (associated with the evaluation point $\textbf{x}$). This time, the geometry $u$ is not represented by its pixelized representation $\textbf{X}$, but the coefficients $\alpha$ coming from the latent space of the VRRAE. In other words, $\tilde{\alpha}$  is the parametrization of $u$.

The thermal problem considered in this work consists in solving the steady-state heat diffusion equation on square 2D plates of size  $4 \times 4$ $m^2$. These geometries are equivalent to the $128 \times 128$ pixel images with which we trained the VRRAE. Each geometry includes four cooling holes (two circles and two squares of equal size) randomly placed within the plate. The boundary conditions are defined such that the outer surface of the solid plate is set at $100^{\circ}\mathrm{C}$, while the cooling holes are kept at $0^{\circ}\mathrm{C}$. The objective is to predict the temperature gradient $\nabla T(u,\mathbf{x})$ 
at any spatial location $\mathbf{x}=(x,y) \in \Omega(u) \subset \mathbb{R}^2$ within the plate. 

Ground-truth solutions are generated using the \textsc{Matlab} PDE Toolbox, which solves the general scalar PDE
\begin{equation}
m \frac{\partial^2 u}{\partial t^2} + d \frac{\partial u}{\partial t} - \nabla \cdot (c \nabla u) + a u = f,
\end{equation}
via a finite element discretization. In our setting, this corresponds to the transient heat diffusion equation
\begin{equation}
\rho C \, \frac{\partial T}{\partial t} - \nabla \cdot \left( k \nabla T \right) = 0,
\end{equation}
with material parameters $\rho = 2700 \,\text{kg/m}^3$, $C = 900 \,\text{J/(kg K)}$, and $k = 237 \,\text{W/(m K)}$. 
Dirichlet boundary conditions are applied: the outer plate boundary is fixed at $100^{\circ}\mathrm{C}$, while the cooling holes are set to $0^{\circ}\mathrm{C}$. The system is simulated until reaching steady state, and the final temperature field is extracted as the reference solution for training.

For this thermal problem described, DeepONet adapts as follows::
\begin{itemize}
  \item Branch net: The input is the 8-dimensional latent vector produced by the trained VRRAE. This vector encodes the geometry of the plate and serves as a compact representation of the input function. The branch net is implemented as a fully connected neural network (FCNN).
  \item Trunk net: The input consists of the 2D spatial coordinates $\mathbf{x}$ where the temperature gradient is to be predicted. The trunk net is likewise implemented as an FCNN.
  \item Output: A scalar value representing the predicted gradient at each queried 2D location for the given plate geometry.
\end{itemize}

This architecture aligns naturally with the structure of our dataset, where each data sample is defined as a triplet $(u, \, \mathbf{x}, \, \nabla T(\mathbf{x}))$. The geometry image ($\mathbf{X}$) is first encoded by the VRRAE into an 8-dimensional latent vector, corresponding to the sampled coefficients $\tilde{\alpha}$ (as described in the VRRAE formulation and illustrated \ref{fig:net_structure}). This latent representation serves as the input to the branch network, while the 2D spatial coordinates $y$ are provided to the trunk network. Their outputs are then combined to produce the final prediction of the temperature gradient at each location.

We adopt the unstacked formulation of DeepONet, where the branch network outputs a vector of $p$ features and the trunk network produces another $p$ -dimensional vector for each coordinate. These vectors are combined via element-wise multiplication and summed to yield the scalar output. The parameter $p$ is a tunable hyperparameter that controls the width of the hidden layers in the subnets. In practice, $p$ is empirically selected based on validation performance and computational considerations. This variant is not only more parameter efficient, but also facilitates faster training compared to the stacked alternative.

A key strength of DeepONet lies in its modular structure, which explicitly separates the representation of input functions from the evaluation points. This architectural design mirrors the prior knowledge that the operator output $G(\tilde{\alpha})(y)$ depends on two independent inputs: $\tilde{\alpha}$ (captured by the branch network) and the location $\textbf{x}$ (encoded by the trunk network). Such alignment between the network structure and the mathematical formulation introduces a strong inductive bias, enabling the model to capture complex operator mappings with improved generalization, even when trained on limited data. Previous studies \cite{Lu_2021} have further demonstrated that this design leads to high-order convergence rates in operator learning tasks.

For training, we frame the problem as a regression task, using the Mean Squared Error (MSE) loss to measure the discrepancy between predicted gradients $G(\tilde{\alpha})(y)$ and the ground truth. This loss is particularly suitable for continuous operator approximation problems, where the aim is to model a nonlinear mapping from geometric configuration and spatial position to a scalar physical quantity.

Finally, the structure of DeepONet lends itself to highly efficient inference, especially when evaluating many geometries over a shared coordinate grid. Since the trunk network depends only on coordinates, it can be evaluated once and reused across all predictions. Then, for each new plate geometry, only the corresponding latent vector must pass through the branch network. The outputs are multiplied with the precomputed trunk output to generate the full prediction. This reuse dramatically reduces computation time during batched inference across large geometry ensembles, making DeepONet particularly advantageous for real-time and many-query scenarios.

\subsection{Model Architecture, Data Generation and Training Details}

As mentioned before, the proposed pipeline consists of two main components: a Variational Rank-Reduction Autoencoder (VRRAE) for compressing geometric information, and a Deep Operator Network (DeepONet) for predicting the temperature gradient field. Both networks are trained independently with separate loss functions and optimization schedules.

\subsubsection*{VRRAE Configuration}
The VRRAE compresses each input geometry into a compact latent representation using a rank-truncated variational bottleneck. The encoder consists of three convolutional layers with channel sizes [32, 64, 128], kernel size 5, stride 2, and padding 1. The decoder mirrors this structure with four transposed convolutional layers of sizes [256, 128, 32, 8], using kernel size 3, stride 2, and padding 1. All layers use ReLU activation, except the final decoder layer, which applies a sigmoid activation to constrain outputs between 0 and 1.

The maximum number of retained singular values is set to $k^* = 8$, which also defines the input dimensionality of the DeepONet’s branch network. During training, the KL divergence term is gradually annealed from zero to a final value of $\beta_{KL} = 0.2$. The network optimizes the Loss function showed in \ref{eqn:loss_vrrae} using Adam with a learning rate of $10^{-4}$, a batch size of 64, and trained for 450,000 steps.

\subsubsection*{DeepONet Configuration}
The DeepONet maps the encoded geometry and spatial coordinates to scalar temperature gradients. We adopt the unstacked DeepONet architecture, where the branch and trunk networks produce $p = 128$ features each, which are combined via element-wise multiplication and linearly projected to the output.

The branch network receives the 8-dimensional latent vector from the VRRAE as input and consists of three fully connected layers with 128 hidden units each. The trunk network takes 2D spatial coordinates as input and consists of four fully connected layers, also with 128 units per layer. ReLU is used as the activation function throughout. The model is trained using the Mean Squared Error (MSE) loss with the Adam optimizer, a learning rate of $10^{-3}$, and a batch size of 10,000 for 200 epochs.

\subsubsection*{Dataset Design}
The dataset comprises $100{,}000$ synthetic binary images of size $128 \times 128$, each representing a square plate with embedded cooling regions. Every plate contains exactly four shapes: two circles and two squares of equal dimensions. The positions of these shapes are assigned randomly within the domain, while ensuring that they do not overlap with the boundaries and that no geometry is repeated across the dataset. White pixels correspond to the solid plate, while black pixels represent cooling holes.

For the DeepONet training, a subset of $5{,}000$ geometries was selected, and their associated steady-state temperature gradient fields were computed using a finite-difference solver implemented in MATLAB. This subset provides paired data consisting of the input geometries and the corresponding physical fields required for operator learning. For each geometry, the aluminum base plate  ($\rho = 2700 \,\text{kg/m}^3$, $C = 900 \,\text{J/(kg·K)}$, $k = 237 \,\text{W/(m·K)}$)  was initialized at $100^{\circ}\text{C}$, with its outer edges held at $20^{\circ}\text{C}$  to mimic thermal contact with an external bath. Internal square and circular inclusions were treated as perfectly cooled regions, with their boundaries fixed at $0^{\circ}\text{C}$. The system was evolved until reaching steady state, and the resulting temperature fields were extracted on a $128 \times 128$ grid. These simulations provide input–output pairs linking geometric configurations to their corresponding stationary thermal fields, which serve as training data for the DeepONet.

Both datasets follow the same partitioning strategy, with $80\%$ of the samples used for training, $10\%$ for validation, and $10\%$ for testing.

\section{Results}

\subsection{VRRAE Training Results}

As shown in Table~\ref{tab:vrrae_results}, the proposed VRRAE achieved a slightly lower reconstruction error (Mean MSE = 0.00820) compared to a standard Autoencoder (Mean MSE = 0.00892), indicating marginal improvement in direct input reconstruction.  

To assess the plausibility of geometries generated from the latent space—either by random sampling or interpolation between known samples—we introduce a custom structural consistency metric. This metric is designed to verify whether the reconstructed images contain a valid number of cooling elements and whether these features occupy a physically plausible area within the image.

Specifically, we count the number of distinct figures detected in the binary mask of the image and compute their combined area as a percentage of the total image size. The generated image is considered valid if the number of detected figures is 4, and if the total occupied area lies within a $5\%$ tolerance of the expected target range, derived from the training dataset.
This evaluation provides an interpretable, domain-specific criterion to verify whether the latent space preserves structural constraints of the geometry. It complements traditional pixel-wise error metrics by enforcing semantic plausibility.

In this regard, the VRRAE exhibited a significantly stronger ability to generate structurally valid geometries. The interpolation metric reached $0.868$ for the VRRAE compared to $0.714$ for the AE, while the random sampling metric achieved $0.741$ for the VRRAE versus $0.567$ for the AE. This metric also reflect the tendency of the AE to produce noisier samples, with bodies that cannot be fully identified as additional cooling points.  These results validate the robustness of the VRRAE against posterior collapse and highlight its ability to learn a coherent and smoothly interpolatable latent space. This property is crucial for integration with the DeepONet, as it ensures that the branch network receives a reliable and low-dimensional representation of the input geometry. For a deeper analysis and comparison of VRRAE with other architectures, we encourage the interested reader to check \cite{mounayer2025rankreductionautoencoders,mounayer2025variational}.

\begin{figure}
  \centering
   \includegraphics[width=\textwidth]{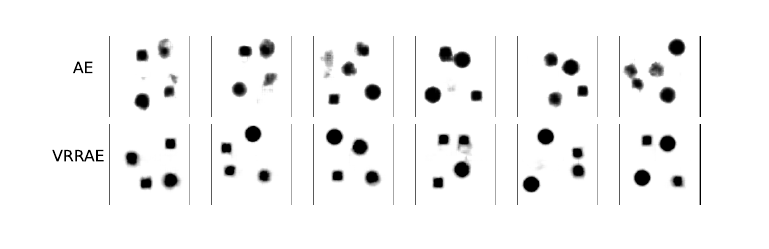}
  \caption{Randomly generated geometrical samples for each one of the selected models.}
  \label{fig:random_geo}
\end{figure}

\begin{figure}
  \centering
  \includegraphics[width=\textwidth]{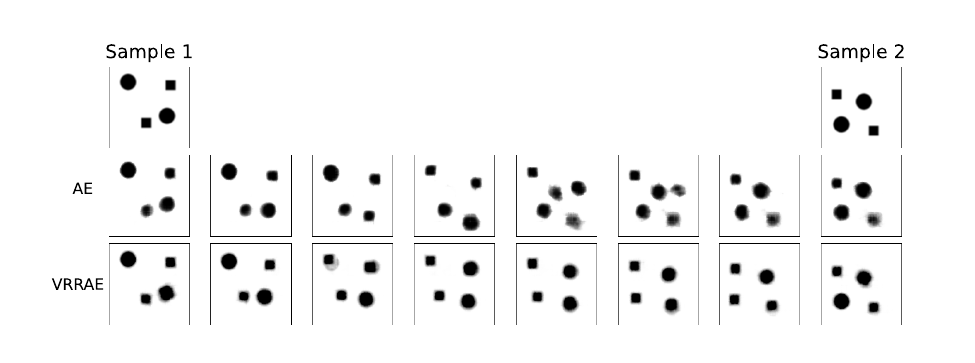}
  \caption{Example of an interpolation (linear, in the latent space) between two samples.}
  \label{fig:inter_geo}
\end{figure}

Finally, it is worth noting that the VRRAE is capable of learning configurations of four components that were not present in the training dataset. For instance, as presented in Figs. \ref{fig:random_geo} and \ref{fig:inter_geo}, not only there are combinations of two circles and two squares far from the original distributions, but also combinations of three squares and one circle, combination that the VRRAE learns to properly interpolate in the manifold consistently based on high quality reduced features.

\begin{table}[h]
\centering
\caption{Comparison between standard Autoencoder (AE) and VRRAE in terms of reconstruction and latent space quality.}
\label{tab:vrrae_results}
\begin{tabular}{lccc}
\toprule
Model & Mean MSE & Inter.  & Rdn.  \\
\midrule
AE     & 0.00892 & 0.714 & 0.567 \\
VRRAE  & \textbf{0.00820} & \textbf{0.868} & \textbf{0.741} \\
\bottomrule
\end{tabular}
\end{table}

\subsection{DeepONet Training Results (2$\times$2 Component Analysis)}

We evaluate a 2$\times$2 design crossing the geometry encoder \{AE, VRRAE\} with the prediction head \{CNN decoder, DeepONet\}. Errors are reported on the test set using three metrics: Mean Squared Error (MSE), Normalized MSE (NMSE, our primary metric), and relative $L_\infty$ error (inf\_nrm). Results (mean $\pm$ standard deviation) are given in Table~\ref{tab:deeponet_results}.
The error metrics are defined as follows. 
Let $\hat{\mathbf{y}} \in \mathbb{R}^N$ denote the predicted field and 
$\mathbf{y} \in \mathbb{R}^N$ the corresponding reference field. 

\begin{align}
\text{MSE} &= \frac{1}{N} \sum_{i=1}^{N} \left( y_i - \hat{y}_i \right)^2, \\[6pt]
\text{NMSE} &= \frac{\sum_{i=1}^{N} \left( y_i - \hat{y}_i \right)^2}
                   {\sum_{i=1}^{N} \left( y_i - \bar{y} \right)^2}, \\[6pt]
\text{inf\_nrm} &= \frac{\| \mathbf{y} - \hat{\mathbf{y}} \|_{\infty}}
                        {\| \mathbf{y} \|_{\infty}}.
\end{align}

\paragraph{Key findings.}
\begin{enumerate}
\item \textbf{Encoder effect.} Upgrading the encoder from AE to VRRAE yields the largest gains: NMSE drops by $38.9\%$ with a CNN head (from $9.30{\times}10^{-7}$ to $5.68{\times}10^{-7}$) and by $21.2\%$ with a DeepONet head (from $7.03{\times}10^{-7}$ to $5.54{\times}10^{-7}$).  
\item \textbf{Head effect.} Replacing a CNN by DeepONet notably helps when the latent space is weaker (AE): NMSE improves by $24.4\%$ (from $9.30{\times}10^{-7} to 7.03{\times}10^{-7}$). With the stronger VRRAE latent, DeepONet still attains the best NMSE/MSE, while the CNN obtains a slightly lower worst-case error (inf\_nrm).  
\item \textbf{Best overall.} \textbf{VRRAE+DeepONet} attains the lowest NMSE and MSE with tighter variability, e.g., $\text{NMSE}=(5.54\pm2.02)\times10^{-7}$ and $\text{MSE}=(3.12\pm1.16)\times10^{-3}$, confirming that a structured latent representation paired with operator learning offers the most accurate and stable predictor.
\end{enumerate}
\begin{table}[h]
\centering
\caption{2$\times$2 component analysis: encoder (AE vs VRRAE) $\times$ head (CNN vs DeepONet). Results as mean $\pm$ std on the test set. Lower is better.}
\label{tab:deeponet_results}
\begin{tabular}{lccc}
\toprule
Model & MSE  & NMSE  & inf\_nrm  \\
\midrule
AE+CNN         & $(5.22 \pm 4.85)\times 10^{-3}$ & $(9.30 \pm 8.76)\times 10^{-7}$ & $0.1150 \pm 0.0447$ \\
AE+DeepONet    & $(3.95 \pm 1.63)\times 10^{-3}$ & $(7.03 \pm 2.82)\times 10^{-7}$ & $0.1060 \pm 0.0168$ \\
VRRAE+CNN      & $(3.17 \pm 3.38)\times 10^{-3}$ & $(5.68 \pm 6.59)\times 10^{-7}$ & $\mathbf{0.0896 \pm 0.0357}$ \\
VRRAE+DeepONet & $\mathbf{(3.12 \pm 1.16)\times 10^{-3}}$ & $\mathbf{(5.54 \pm 2.02)\times 10^{-7}}$ & $0.0942 \pm 0.0138$ \\
\bottomrule
\end{tabular}
\end{table}

\begin{figure}
  \centering
  \includegraphics[width=\textwidth]{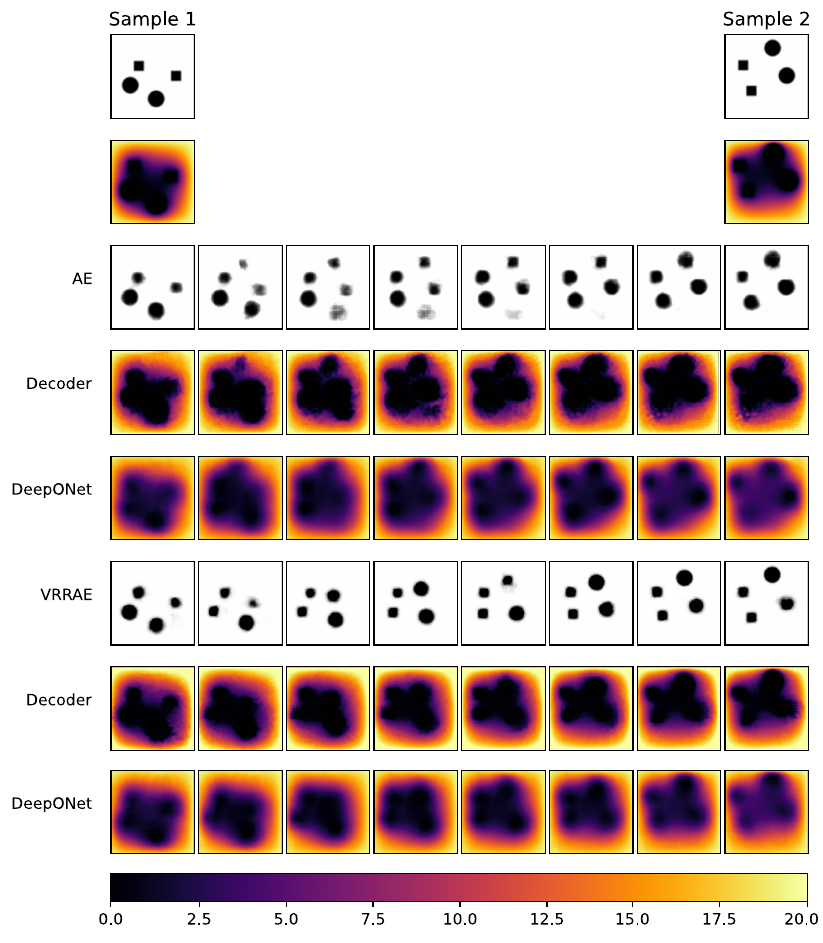}
  \caption{Linear interpolation in the latent space between two geometries (Sample~1 and Sample~2). 
The first row shows the input geometries, followed by their corresponding 
steady-state thermal fields. 
Intermediate rows illustrate reconstructions and interpolated fields obtained with 
the Autoencoder (AE), Variational Rank-Reduced Autoencoder (VRRAE), and DeepONet models. 
The colorbar indicates temperature in $^{\circ}$C.}
  \label{fig:inter_heat}
\end{figure}

An important advantage of the proposed VRRAE lies in the quality and consistency of its latent space. As illustrated in Fig.~\ref{fig:inter_heat}, the VRRAE encodes geometries into a representation that is substantially more structured and contains far fewer ``holes'' than that of a standard AE. As a consequence, decoded samples or latent interpolations are much less likely to produce non-physical outcomes, such as an incorrect number of cooling sources (e.g., five elements) that were never present in the training data. Although such spurious cases cannot be completely eliminated, their occurrence is drastically reduced compared to the AE. This highlights that the improvement brought by the VRRAE is not only reflected in lower numerical errors, but also in a latent space that is more robust, physically plausible, and suitable for downstream operator learning with the DeepONet. 

\section{Conclusions}

In this work, we addressed the challenge of developing a fast and accurate thermal solver for temperature fields over complex geometries, a task that is traditionally computationally expensive. Existing deep learning approaches, such as autoencoders (AEs) and variational autoencoders (VAEs), often suffer from unstructured latent spaces or ``holes'', which limit their effectiveness for design exploration and the generation of physically consistent solutions.  

To overcome these limitations, we proposed a hybrid framework that combines Variational Rank-Reduction Autoencoders (VRRAEs) with Deep Operator Networks (DeepONets). This approach enables reduced-order parametrization of design domains and efficient prediction of temperature gradients for variable geometries.  

Our main contributions can be summarized as follows:

\begin{itemize}
    \item \textbf{Structured Latent Geometry with VRRAEs:} By incorporating truncated singular value decomposition (SVD) into the latent space and setting the latent mean function to the identity, the VRRAE learns a structured, interpretable, and continuous representation of geometry. This mitigates the ``holes'' problem and improves robustness against posterior collapse, while maintaining reconstruction quality without requiring additional loss terms beyond the standard KL divergence.  
    
    \item \textbf{Efficient Operator Prediction with DeepONets:} The DeepONet leverages the compact 8-dimensional latent vector provided by the VRRAE in its branch network and the spatial coordinates in its trunk network. This modular structure is well-suited for nonlinear operator approximation and enables accurate prediction of temperature gradients. A key advantage is the efficiency in inference: while Abaqus requires an average of 0.273\,s per sample to compute gradients, our DeepONet achieves 0.0026\,s per sample, offering more than two orders of magnitude speedup.  
    
    \item \textbf{Performance of the Hybrid Approach:} Experimental results show that the VRRAE+DeepONet achieves the highest accuracy in temperature gradient prediction, with the lowest MSE and NMSE values and reduced variability on the test set. The 2D component analysis further highlights that upgrading from a standard AE to a VRRAE yields the largest performance gains, underscoring the importance of structured latent representations.  
\end{itemize}

This study demonstrates the potential of combining generative models and operator networks for thermal design and, more broadly, for engineering applications requiring efficient exploration of solution spaces.  

\section*{Acknowledgements}

This work was supported by the Spanish Ministry of Science and Innovation,
AEI/10.13039/501100011033, through Grant number PID2023-
147373OB-I00, and by the Ministry for Digital Transformation and the Civil
Service, through the ENIA 2022 Chairs for the creation of university-industry
chairs in AI, through Grant TSI-100930-2023-1.

This research is also part of the DesCartes programme and is supported by the National Research Foundation, Prime Minister Office, Singapore under its Campus for Research Excellence and Technological Enterprise (CREATE) programme.

This material is also based upon work supported in part by the Army
Research Laboratory and the Army Research Office under contract/grant
number W911NF2210271.

The authors also acknowledge the support of Keysight Technologies through the chairs at the University of Zaragoza and at ENSAM Institute of Technology, and SKF Magnetic Mechatronics at ENSAM.

B. Moya acknowledges support from the French government, managed by the National Research Agency (ANR), under the CPJ ITTI.

\bibliography{references} 
\bibliographystyle{unsrt}

\end{document}